\relax
\documentclass[letterpaper]{article} 
\usepackage{aaai22}  
\usepackage{times}  
\usepackage{helvet}  
\usepackage{courier}  
\usepackage[hyphens]{url}  
\usepackage{graphicx} 
\usepackage{natbib}  
\usepackage{caption} 
\DeclareCaptionStyle{ruled}{labelfont=normalfont,labelsep=colon,strut=off} 
\usepackage{algorithm}
\usepackage{algorithmic}

\usepackage{amsmath}
\usepackage{epsfig}
\usepackage{tabularx}
\usepackage{multicol}
\usepackage{multirow}
\usepackage{subfigure}

\usepackage[switch]{lineno}

\newcommand\etc{\textit{etc.}}
\newcommand\etals{\textit{et al}~}
\usepackage{newfloat}
\usepackage{listings}
\floatstyle{ruled}
\newfloat{listing}{tb}{lst}{}
\floatname{listing}{Listing}
\title{ Multi-Scale Temporal Difference Transformer for Video-Text Retrieval}
\author {
   Ni  Wang,\textsuperscript{\rm 1}
    Dongliang Liao, \textsuperscript{\rm 2}
    Xing Xu\textsuperscript{\rm 1}
}
\affiliations {
    \textsuperscript{\rm 1} University of Electronic Science and Technology of China\\
    \textsuperscript{\rm 2}  WXG Tencent\\
    wangni1430@gmail.com
}
\usepackage{bibentry}

\begin{document}
\maketitle

\begin{abstract}
Currently, in the field of video-text retrieval, there are many transformer-based methods. 
Most of them usually stack frame features and regrade frames as tokens, then use transformers for video temporal modeling.
However, they commonly neglect the inferior ability of the transformer modeling local temporal information.
To tackle this problem, we propose a transformer variant named $\textit{Multi-Scale Temporal Difference Transformer}$ (MSTDT). MSTDT mainly addresses the defects of the traditional transformer which has limited ability to capture local temporal information.
Besides, in order to better model the detailed dynamic information, we make use of the difference feature between frames, which practically reflects the dynamic movement of a video.
We extract the inter-frame difference feature and integrate the difference and frame feature by the multi-scale temporal transformer.
In general, our proposed MSTDT consists of a short-term multi-scale temporal difference transformer and a long-term temporal transformer.
The former focuses on modeling local temporal information, the latter aims at modeling global temporal information.
At last, we propose a new loss to narrow the distance of similar samples.
Extensive experiments show that backbone, such as CLIP, with MSTDT has attained a new state-of-the-art result.
\end{abstract}

\section{Introduction}
Video-text retrieval is a challenging domain, which needs the model to learn the information of two modalities. Entering the era of information explosion, a large number of pre-training models obtained better results through massive training data. Some methods are pre-trained on videos \cite{ht, mmt, t2vlad}, these models definitely show a great improvement in the video-text retrieval task. 
There are some models that also perform excellently through image-text pre-training \cite{clipbert}.
A pioneering work is Contrastive Language-Image Pre-training (CLIP) \cite{clip}, which trained with 400 million image-text pairs and showed a remarkable ability in learning fine-grained visual concepts for images.
CLIP performed so well on the image-text task that many works transitioned CLIP from image-text to video-text tasks. \citet{clips} directly applied CLIP to the video-text retrieval task.
Compared with many video pre-training works, this direct application for zero-shot prediction has obtained a comparable result, even not making use of sufficient video temporal information. 

 \begin{figure}[!tb]
    \centering
    \includegraphics[width = 0.99\columnwidth]{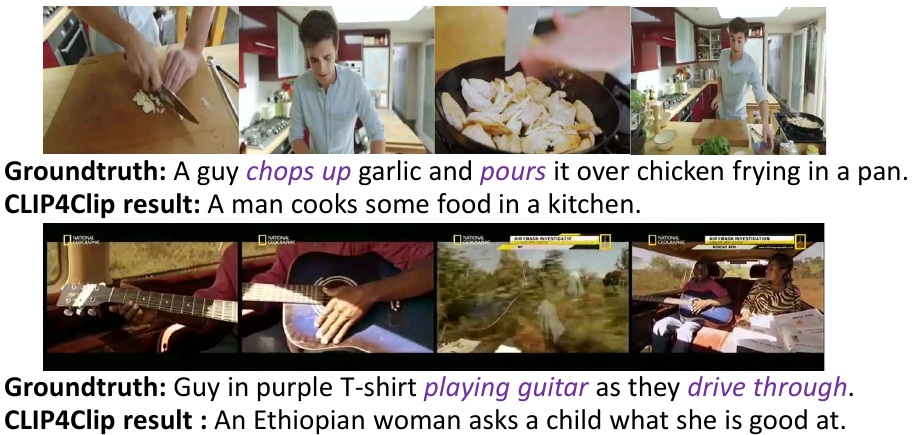}

    \caption{Failed cases of CLIP4Clip. CLIP4Clip can correctly model the context of the entire video, such as ``cook in kitchen” and ``woman asks a man", but it is not good at short-term fine-grained actions and subtle scenes, such as actions colored with purple.}
    \label{example}
\end{figure}

Better modeling video temporal information has become a crucial issue to improve the effectiveness of CLIP in video-text retrieval task. CLIP4Clip \cite{clip4clip} added a  4-layer transformer encoder after visual transformer (ViT) \cite{vit} of the CLIP to learn temporal information.
Its success indicated that exploiting temporal information can indeed effectively improve the performance of CLIP.

However, there are some problems of the transformer to be tackled:
transformer does not do well in local temporal information modeling. It mainly focuses on global interaction, so it tends to pay too much attention to global information while weakening neighboring information. \citet{cnn_tf_battle} also proved the limitations of transformers in local modeling through sufficient experiments. 
Fig \ref{example} shows some failed cases of CLIP4Clip to show the drawback of traditional transformer.
It is evident that CLIP4Clip has effective long-term temporal inference ability and background representation. 
However, it cannot make precise representations of detailed entities and actions. 
\textbf{How to make transformer have local modeling ability } is one critical problem to be solved in this paper.

To tackle the above issue, we propose a transformer variant called \textit{Multi-Scale Temporal Transformer} (MSTT).
It explicitly splits a sequence to several multi-scale short-term subsets.
It is easy to understand that various short-term subset also reflects various local parts of the full sequence.

Current methods are pay more attention to the background in a video, and the detailed actions are always ignored. Understanding dynamic information, such as actions and scene transitions, is a complex task, since dynamic information is constantly changing in the temporal dimension. \textbf{How to improve the detailed dynamic information modeling ability of the model} is another problem to be solved in this paper.

The difference feature, which brought a great improvement in several computer vision tasks \cite{lbp, cdc, dc_cdc, cdc_rsp}, has remarkable representation ability for invariant fine-grained features in diverse environments. It reflects the difference between frames.
In order to further enhance the model's ability to understand detailed temporal information, such as actions and scene transitions, we incorporate the difference feature.
On the basis of \textit{Multi-Scale Temporal Transformer}, we further propose \textit{Multi-Scale Temporal Difference Transformer} (MSTDT).
We extract the difference feature and fuse the frame and difference features through the proposed MSTDT.


The conventional loss functions usually focus on how to make the ground truth closer. It is also essential to narrow the distance of similar sample. Supposing video1 and caption1 are ground truth pair, and video2 has similar semantics to video1. We think the similarity score of video1 and text1, video1 and text2, video1 and video2 should all be high. \textbf{How to narrow the distance of similar samples} is the last problem to be solved in our paper. \textit{Binary Similarity Loss} is proposed to ensure the distance between similar samples from the level of distribution alignment. It allows samples with the same semantics to get closer distances, so that the model can learn a more powerful semantic space.

In general, our contributions can be summarized as follows:
\begin{itemize}
    \item We propose the Multi-Scale Temporal Transformer, which not only focuses on global temporal modeling but also excels at local temporal modeling. Effective use of multi-scale also helps model with better modeling local information in different receptive fields.
    \item To further enhance the detailed dynamic information modeling ability, we incorporate the difference feature. By combining the difference and frame features, our method performs better in detailed temporal modeling.
    \item We propose the binary similarity loss to ensure the distance of similar samples in common space for video-text similarity prediction, which brings a sensible improvement to already very strong results.
    \item Extensive experiments show the effectiveness of our proposed method and its new state-of-the-art performance on video-text retrieval task.

\end{itemize}

\section{Related Work}
\subsubsection{Image-Text Representation}
With the rise of big data and the appearance of transformers \cite{att_is_all}, more and more scholars pay attention to pre-training methods.A lot effective models have come out. 
\citet{vilbert} proposed ViLBERT, which is composed of two transformer encoders. The image and text are encoded separately by each transformer encoder and finally they are fused through a new common attention layer. 
\citet{imagebert} proposed a new visual language pre-trained model named ImageBERT, as well as a large weakly supervised image-text dataset LAIT. Through multi-stage training and several pre-training tasks, ImageBERT has made a great improvement in image-text task. 
Recently, OpenAI proposed CLIP \cite{clip}, which consists of visual transformer (ViT) \cite{vit} encoder on the video side and text transformer encoder on the text side. CLIP showes an excellent ability in learning image representations from linguistic supervision. These image pre-training methods had shown the powerful representation of fine-grained visual concepts and spatial information. Therefore, how to transfer these significant image-text features to video-text tasks and effectively integrate video temporal information has become an important issue of our concerns.
\subsubsection{Video-Text Retrieval}
Recently, mainstream video-text retrieval methods are pre-training transformer-based models.
\citet{mmt} presented a multi-modal transformer(MMT) to jointly encode the different modalities. MMT allows different modalities to attend to the others on the video side.
ClipBERT \cite{clipbert} enabled the video language tasks to achieve affordable end-to-end learning by using sparse sampling, which can reduce a lot of overheads for long videos.
\citet{clips} applied CLIP to video text retrieval tasks without extra modification. Its great result is comparable with many video pre-training methods.
Luo \etals proposed an improved model named CLIP4Clip \cite{clip4clip}, which extended CLIP with post-pretraining on video dataset and transferred the image-language features to video-language retrieval in an end-to-end manner.
CLIP2Video \cite{clip2video} added two temporal blocks to explore the temporal relation of the video. 
These two models, exploring video temporal information, performed better than the basic CLIP model. 
Different from the above two methods, we propose a new method to model video temporal information in two aspects: short-term and long-term. We further adopt difference information to enhance the motion feature.

 \begin{figure*}[!htb]
    \centering
    \subfigure[MSTDT] 
    {
    \includegraphics[width = 0.47\textwidth]{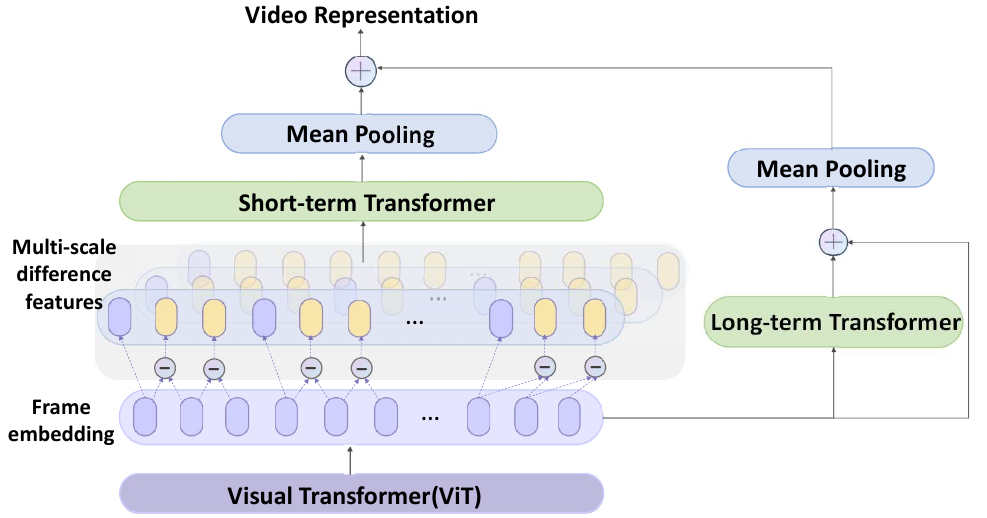}
}
    \subfigure[CLIP-MSTDT]
    {
  \includegraphics[width = 0.5\textwidth]{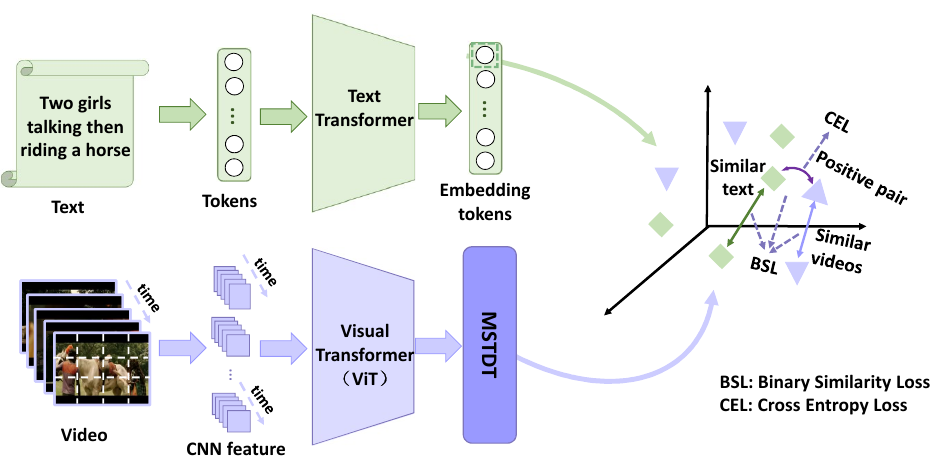}
    }
    \caption{Overview of the proposed  Multi-Scale Temporal Difference Transformer (MSTDT) and CLIP with Multi-Scale Temporal Difference Transformer (CLIP-MSTDT).}
    \label{framework}
\end{figure*}


\section{Proposed Method}
In this section, we firstly give some problem formulations. Secondly, we specify our method \textit{Multi-Scale Temporal Difference Transformer}(MSTDT), Fig. \ref{framework} (a) illustrates the framework of MSTDT. Thirdly, we instantiate our MSTDT with the CLIP backbone. Fig. \ref{framework} (b) illustrates the overview of CLIP with MSTDT. For captions, we take the output of a text transformer encoder as final caption embedding. For videos, we employ MSTDT to model short-term and long-term temporal on the basis of visual transformer (ViT) encoder. Finally, we specify a new loss named binary similarity loss, which mainly considers the distance of similar samples.
\subsection{Problem Formulation}
Let $V$ and $C$ denote a video and a video caption, respectively. A video $V$ is affiliated with at least one caption $C$.
A video can be represented as a sequence of frames $V$ = $\left\{\textit{\textbf{v}}_1, \textit{\textbf{v}}_2, \ldots,\textit{\textbf{v}}_i, \ldots,\textit{\textbf{v}}_n\right\}$, where $\textit{\textbf{v}}_i$ denotes the feature vector of the $i$-th frame, $n$ is video length. A caption can be represented as a sequence of words $C$=$\left\{\textit{\textbf{c}}_1, \textit{\textbf{c}}_2, \ldots, \textit{\textbf{c}}_j, \ldots,\textit{\textbf{c}}_l\right\}$, where $\textit{\textbf{c}}_j$ denotes the feature vector of the $j$-th word, $l$ is caption length. Given a video set and its caption set, the goal of video-text retrieval task is to identify the desired video(caption) to be close to the semantics of query caption(video).

\subsection{Multi-Scale Temporal Difference Transformer}
As Fig. \ref{framework} (a) shows, MSTDT consists of two blocks: a short-term multi-scale temporal difference transformer and a long-term temporal transformer. The short-term module is targeted at modeling local temporal information, which is a defect of the traditional transformer. The long-term is aimed at modeling global temporal information and static information. We will introduce the MSTDT in a progressive way.

\subsubsection{Short-term Temporal Transformer}
We firstly divide a video into several frame subsets. Here we use $k$ denotes the scale of each frame subsets, so we can obtain $m$ = $\lfloor \frac{n}{k} \rfloor$ subsets. As our goals are modeling local temporal information and improving the interaction rate of the conventional transformer. Each frame subset represents a part local information of a video.
Given a $k$-scale divided video, it is composed of $m$ frame subsets. Each subset is represented as $\textit{S}^k$ = $[\textit{\textbf{v}}_1, \textit{\textbf{v}}_2, \ldots, \textit{\textbf{v}}_{k}]$, we compute the short-term temporal feature $\textit{\textbf{s}}^k$as follows:
\begin{equation}
   \textit{\textbf{s}}^k =\frac{1}{m} \frac{1}{k}\sum_{i=1}^{m} \sum_{t=1}^{k} mask(\textit{\textbf{v}}_t^i)tf(\textit{\textbf{v}}_t^i),
\end{equation}
where $k$ is the scale of each frame set,
$\textit{\textbf{v}}_t^i$ denotes the $t$-th feature of $i$-th frame set. $tf$ is the transformer encoder, $mask(\textit{\textbf{v}}_t^i)$ is the mask generate function, where $mask(\textit{\textbf{v}}_t^i)$ = 0 if $\textit{\textbf{v}}_t^i$ is a zero-padding frame.

\subsubsection{Short-term Multi-Scale Temporal Transformer}
Referring to the idea that different kernel sizes in CNN generate different receptive fields, we form the Multi-Scale Temporal Transformer (MSTT) with various $k$, such as $k$= 3, 4, 6. We use $K$ to represent the scale list. Then we take the mean pooling fusion on multi-scale short-term embedding features, denoted as:
\begin{equation}
  \textit{\textbf{s}} =\frac{1}{|K|} \sum_{k \in K} \textit{\textbf{s}}^k,
\end{equation}

\subsubsection{Short-term Multi-Scale Temporal Difference Transformer}
Difference information more clearly reflects the motion between frames, so we adopt difference features for better detailed dynamic information. We stack the extract temporal difference $\textit{D}_k$ = $[\textit{\textbf{d}}_1, \textit{\textbf{d}}_2, \ldots, \textit{\textbf{d}}_{k}]$, where $\textit{\textbf{d}}_i$ = $\textit{\textbf{v}}_{i+1}-\textit{\textbf{v}}_i$. Specially, for the last non-padding frame $\textit{\textbf{v}}_t$, $\textit{\textbf{d}}_t$ = $\textit{\textbf{v}}_1-\textit{\textbf{v}}_t$, for padding frames, the related difference features equal zero. 
However, just depending on difference features can't correctly identify actions, as different actions may have the same trajectory.
So we add the first frame feature of each subset for visual guiding. The guiding frame and difference feature are represented as $\textit{D}_k$ = $[\textit{\textbf{d}}_0, \textit{\textbf{d}}_1, \textit{\textbf{d}}_2, \ldots, \textit{\textbf{d}}_{k}]$, $\textit{\textbf{d}}_0$ denotes the guiding frame. We also learn the position embedding of the guiding frame and each difference feature $\textit{
P}_k$ = $[\textit{\textbf{p}}_0, \textit{\textbf{p}}_1, \textit{\textbf{p}}_2, \ldots, \textit{\textbf{p}}_{k}]$. Then we add each item of $\textit{D}_k$ and $\textit{P}_k$. 
Based on above representation, The final short-term embedding feature $\textit{\textbf{s}}$ is as follows:
\begin{equation}
  \textit{\textbf{s}} =\frac{1}{|K|} \frac{1}{m} \frac{1}{k}\sum_{k \in K}\sum_{i=1}^{m} \sum_{t=0}^{k} mask(\textit{\textbf{p}}_t^i + \textit{\textbf{d}}_t^i) tf(\textit{\textbf{p}}_t^i + \textit{\textbf{d}}_t^i),
\end{equation}
$mask(\textit{\textbf{p}}_t^i + \textit{\textbf{d}}_t^i)$ = 1 if $\textit{\textbf{d}}_t^i$ represent frame feature in the subset. For difference feature, $mask(\textit{\textbf{p}}_t^i + \textit{\textbf{d}}_t^i)$ = 1 if neither $\textit{\textbf{v}}_t^i$ or $\textit{\textbf{v}}_{t+1}^i$ is zero-padding frame, else $mask(\textit{\textbf{p}}_t^i + \textit{\textbf{d}}_t^i)$ = 0. 

\subsubsection{Long-term Temporal Transformer}
The long-term module mainly focuses on global temporal information and static information fusion.
Follow the traditional transformer with slightly different settings in the text task. We don't add another [CLS] token or regard it as the final representation feature. The backbone module, such as ViT of CLIP, already has powerful characterization capabilities. CLIP4Clip has approved the average feature of ViT has gained an excellent result. So we combine the original embedding features and transformer modeled embedding features through a residual connection. Finally, we take the average of added embedding features as long-term embedding features \textit{\textbf{l}}. The module is denoted as follows: 

\begin{equation}
   \textit{\textbf{l}} =\frac{1}{n} \sum_{i=1}^{n}  mask(\textit{\textbf{v}}_i)(\textit{\textbf{v}}_i + tf(\textit{\textbf{v}}_i)) ,
\end{equation}
where $n$ is the length of video. Finally, we combine short-term embedding feature \textit{\textbf{s}} and long-term embedding feature \textit{\textbf{l}}. The final video embedding $V_f$ is represented as:
\begin{equation}
   V_f = \alpha*\textit{\textbf{s}} +(1-\alpha)*\textit{\textbf{l}},
\end{equation}
$\alpha$ is a temporal trade-off rate. 

\subsection{Exemplar: CLIP with MSTDT}
Fig. \ref{framework} (b) illustrates the overview of CLIP with MSTDT. CLIP \cite{clip} is a cross-modal general model, which is pre-trained with 400 million image-text pairs. As it is pre-trained on such a massive data set, its embedding features are versatile for all cross-modal tasks. Text encoder is 12-layer 512-wide transformers with 8 attention heads. Following CLIP, the embedding features from the highest layer of the transformer at the [CLS] token are treated as the feature representation of the captions. For a caption $C$, we denote the ultimate representation as $C_f$. The video encoder consists of three parts: convolution layers, 12-layer 512-wide visual transformer encoder with 8 attention heads, and MSTDT. we extract patches' features of each frame $\textit{\textbf{v}}_i$ through the convolutional layer. Then we explore the spatial information of each frame by visual transformer encoder. After that, we explore the temporal information of all frames by MSTDT. For a video $V$, we denote the final representation as $V_f$. We map $C_f$ and $V_f$ into a common space and measure their similarity through cross entropy loss and binary similarity loss.

\subsection{Loss function}
The conventional loss functions pay more attention to how to make the ground truth closer. While it is also essential to narrow the distance of similar videos, similar text, and similar ground truth pairs. We provide an example in Fig \ref{bml_fig}. Rectangles denote video features, triangles represent text features. The ground truth pairs are given the same color. The more similar the color of the ground truth pairs, the more similar the semantics. In the left figure of Fig \ref{bml_fig}, we can see although these ground truth pairs are closer enough, the similar samples might separated (marked with green and purple arrows). As the loss like cross entropy just narrow the distance of ground truth. During this process, the samples in similar semantics may not be able to maintain a close distance. Therefore, we design binary similarity loss to ensure the distance of similar samples, like the right figure of Fig \ref{bml_fig}.

For each batch, video features and caption features are denoted as $x^v$ = $( x_1^v, x_2^v, \cdots, x_b^v )$ and $x^c$ = $( x_1^c, x_2^c, \ldots,x_b^c)$. Here $b$ indicates the mini-batch size. To measure the video-text similarity matrix $sim(x^v; x^c)$, we use popular cosine similarity between $x^v$ and $x^v$. Similarly, we also calculated $sim(x^v; x^v)$ and $sim(x^c; x^c)$. 
When doing video-to-text retrieval, we softmax $sim(x^v; x^c)$ and $sim(x^v; x^v)$ by row.
When doing video-to-text retrieval, we softmax $sim(x^v; x^c)$ and $sim(x^c; x^c)$ by column, respectively. 
The diagonal elements of the matrix are the scores of ground truth pairs.
To avoid the effect of extremely high diagonal values, we mask the diagonal values of the matrix. Then we compute the KL divergence of each two matrices as follows:
 \begin{figure}[!tb]
    \centering
    \includegraphics[width = 0.49\textwidth]{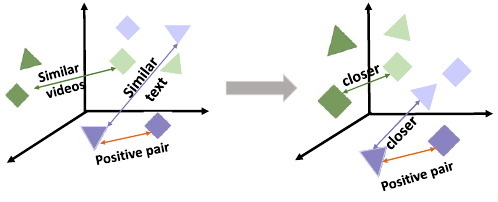}
    \caption{Examples illustrating the purpose of binary similarity loss.}
    \label{bml_fig}
\end{figure}

\begin{equation}
    \begin{aligned}
        \mathcal{L}_{bs}  =  &KL((sim(x^v, x^c)), (sim(x^v; x^v))) \\
             + &KL((sim(x^v; x^c)), (sim(x^c;x^c)))
    \end{aligned}
\end{equation}
\noindent Binary similarity loss measures the distance between similar samples, we use symmetric cross entropy loss to ensure the distance between ground truth pair, denoted as:
\begin{equation}
    \begin{aligned}
        \mathcal{L}_{ce}  =  &-\frac{1}{b}\sum^{b}_{i=1} log \frac{exp(sim(x_i^v, x_i^c))}{\sum^{b}_{j=1}exp(sim(x_i^v, x_j^c))} \\ 
        & -\frac{1}{b}\sum^{b}_{i=1} log \frac{exp(sim(x_i^c, x_i^v))}{\sum^{b}_{j=1}exp(sim(x_i^c, x_j^v))}
    \end{aligned}
\end{equation}

The final loss is composed of above two losses, $\beta$ is a loss trade-off rate.
\begin{equation}
    \begin{aligned}
        \mathcal{L}  = \beta \mathcal{L}_{bl} + (1- \beta) \mathcal{L}_{ce}
    \end{aligned}
\end{equation}

\section{Experiment}
\subsection{Experimental Settings}
\subsubsection{Datasets}
\begin{itemize}
    \item The MSR-VTT \cite{msrvtt} dataset contains 10,000 video clips and each video comes with 20 sentence descriptions. 
    We use two types of data splits, ‘Training-7K’ and ‘Training-9K’, to compare with baselines. The ‘Training-7K’ follows the data splits from \cite{howto100} and the ‘Training-9K’ follows the data splits from \cite{mmt}. The test data in both splits is ‘test 1k-A’, which contains 1,000 clip text pairs following JSFusion \cite{jsfusion}.
    \item The MSVD \cite{msvd} contains 1970 Youtube clips, with approximately 40 associated sentences in English for each video. MSVD is split into 1,200 videos for training, 100 videos for validation, and 670 videos for testing.
    \item The TGIF \cite{tgif} contains 100K animated GIFs and 120K sentences describing the visual content of the animated GIFs. We split 78,654 videos for training and 11,330 videos for testing. Each video is annotated with one description when training and three descriptions when testing. 
\end{itemize}

\subsubsection{Baselines}
\begin{itemize}
    \item \textbf{Non-pretraining models:} JSFusion \cite{jsfusion}, DE \cite{Dual_Encoding}, CE \cite{ce}, ARRN \cite{arrn}, PVSE \cite{PVSE}, HGR\cite{HGR}.
    \item \textbf{Video pre-training models:}  HT-pretrained \cite{ht}, MMT-PT(MMT-pretrained)  \cite{mmt},  MDMMT \cite{mdmmt},
	SupportSet \cite{support_set}, 
    \item \textbf{CLIP based models:}  CLIP\cite{clips}, CLIP4Clip \cite{clip4clip} and CLIP2Video \cite{clip2video}.
\end{itemize}

\subsubsection{Evaluation Metric.}
Following the convention in video-text retrieval, we report rank-based performance metrics, including $R@k$, Median Rank ($MedR$), Mean Rank($MeanR$). $R@k$ ($k = 1, 5, 10$) is the percentage of queries for which the correct item is retrieved among the top $k$ results. The $MedR$ measures the median rank of correct items. Correspondingly, $MeanR$ is the mean rank of correct items. Higher $R@k$, lower $MedR$ and $MeanR$ indicate better performance. $Rsum$ is the sum of $R@k$ to reflect the overall performance.

\subsubsection{Implementation Details}
Following the experiment settings from CLIP4Clip \cite{clip4clip}, we sparsely and uniformly sample 12 frames per video. Then we crop each frame to a 224×224 region. the max length of a caption is 12. The feature dimensions of video and captions are 512. 
We initialize the backbone with pre-training weights of ViT-B/32 \cite{clip}. 
We optimize our model with the Adam optimizer \cite{adma}, with a initial learning rate of 1e-7 for backbone and 4e-4 for MSTDT module. We decay the learning rate using a cosine schedule.
We train the CLIP4Clip model with the source code supplied by \citet{clip4clip}. We implement the CLIP2Video model with the help of evaluation code released by \citet{clip2video}. In order to assure the fairness of the comparison, CLIP4Clip, CLIP2Video and MSTDT use the same sampling strategy. We run these three models at least three times with different seeds and take the average result as the report result. 
The batch size is 128 and running 5 epochs. 
All fine-tuning experiments are carried out on 8 NVIDIA Tesla V100 GPUs. Other baseline results are derived from the original paper.

\subsection{Comparison with State-of-the-Art}
Table \ref{tab:results_msrvtt9k} summarizes the performance comparison on MSR-VTT 9k/7k, MSVD, TGIF and MSR-VTT. 
It can be concluded that our approach significantly outperforms all the baselines on MSR-VTT 9k, MSVD, and TGIF. Compared with video pre-training methods, such as MMT-pretrianed, MDMMT \etc, CLIP-based models are outperforming. Benefit from the large-scale image-text pre-training, CLIP indeed has exceptional spatial modeling ability. Comparing with the results of other CLIP-based methods, our model performs better. We could conclude that considering short-term temporal information and long-term temporal information jointly explicitly helps with model performance. 
In particular, CLIP4Clip and CLIP2Video just consider long-term temporal information. But our method further explores temporal information in the short-term. Some fine actions ignored by those two methods can be accurately captured by our model. The long-term module and the short-term module can effectively compensate for each other's defects, so our method could obtain a state-of-the-art experimental result.

In Table \ref{tab:results_msrvtt9k}, we can see our proposed method gains a large promotion gap when training in MSR-VTT 9K. While when training in MSR-VTT 7K, the improvement is not as apparent as MSR-VTT 9K. As the MSTDT in default contains 12 transformer layers (every scale contains 4 layers), it is hard for MSTDT to learn optimal parameters with a small-scale training data. MSVD's training data is also a small-scale dataset, so the improvement of MSTDT is limited. When we set every scale containing 1 transformer layer (results are recorded in supplementary materials), it performs better than the default setting. This result demonstrates that when training on a smaller dataset, it's better to decrease the number of transformer layers. The result on the TGIF dataset illustrates that our method still works well on short gifs animations. This phenomenon shows our method is not only suitable for long videos, but also for short videos.
\begin{table*}[!htb]
\small
\centering
\setlength\tabcolsep{3pt}
\begin{tabular}{ccc ccccc cccccc}
	\hline
	\multicolumn{1}{c}{\multirow{2}{*}{Data}} &\multicolumn{1}{c}{\multirow{2}{*}{Methods}} & \multicolumn{1}{c}{\multirow{2}{*}{Year}} & \multicolumn{5}{c}{Text-to-Video} & \multicolumn{6}{c}{Video-to-Text} \\
	  & & & R@1 & R@5 & R@10 &MedR  & MeanR & R@1 & R@5 & R@10 & MedR & MeanR &Rsum\\
	\hline 
    \multicolumn{1}{c}{\multirow{10}{*}{\rotatebox{270}{MST-VTT 9K}}} 
    &JSFusion 	&2018
	    &10.2 &31.2 &43.2 &13 &- &- &- &- &- &- &- \\
    &CE &2019
        &20.9 &48.8 &62.4 &6 &28.2   &20.6 &50.3 &64.0 &5 &25.1 &267.0\\
    &HT-PT &2019
        &14.9 &40.2 &52.8 &9 &- &-  &- &- &- &- &-\\
    &MMT-PT &2020
        &26.6 &57.1 &69.6 &4 &24.0  &27.0 &57.5 &69.7 &4 &21.3 &307.5\\
    &SupportSet  &2021 
        &27.4 &56.3 &67.7 &3 &- &26.6 &55.1 &67.5 &3 &- &300.6\\
    &CLIP  &2021 
        &31.2 &53.7 &64.2 &4 &-  &27.2 &51.7 &62.6 &5 &- &290.6\\
    &MDMMT &2021
        &38.9 &69.0 &79.7 &\textbf{2} &16.5  &- &- &- &- &- &-  \\
	&CLIP4Clip &2021
	    &42.3  &72.3  &82.0 &\textbf{2} &14.5  &41.3 &71.0 &81.7 &\textbf{2} &12.4 &391.5\\
    &CLIP2Video &2021
	    &43.6 &71.5 &81.2 &\textbf{2} &14.0  &42.7 &71.7 &82.0 &\textbf{2}  &10.6 &392.7\\
	&\textbf{CLIP-MSTDT (Ours)} &2021
	    &\textbf{44.8} &\textbf{73.3} &\textbf{82.8} &\textbf{2} &\textbf{13.7}
	    &\textbf{43.0} &\textbf{72.4} &\textbf{82.1} &\textbf{2} &\textbf{10.3} &\textbf{398.4}\\
	\hline 
	\multicolumn{1}{c}{\multirow{3}{*}{\rotatebox{270}{7K}}}
    &CLIP4Clip &2021
	    &41.6 &69.5 &79.6 &\textbf{2} &16.7 &40.0 &69.2 &78.5 &\textbf{2} & 13.4 &379.4\\
	&CLIP2Video &2021
	    &41.7 &69.9 &\textbf{80.1} &\textbf{2} &\textbf{15.6} &\textbf{42.5} &\textbf{69.7} &\textbf{79.7} &\textbf{2} & \textbf{11.6} &\textbf{383.6}\\
	&\textbf{CLIP-MSTDT (Ours)} &2021
	     &\textbf{42.8}  &\textbf{70.9}  &79.1  &\textbf{2}  &16.3 &42.0  &69.0  &79.5 &\textbf{2} &12.3 &383.3\\
	\hline 
    \multicolumn{1}{c}{\multirow{6}{*}{\rotatebox{270}{MSVD}}} 
    &DE &2019
	    & 12.7 &34.5 & 46.4 & 13 & -  & 16.1 & 32.1 & 41.5 & 17 & - &183.3 \\

    &ARRN &2021
	    & 16.1 &39.7 &54.1 &9  &-   & 19.6 & 41.9 & 51.1 & 8 & - &222.5 \\
	&SupportSet   &2021
	&28.4 &60.0 &72.9 &4 &- &- &- &- &- &- &- \\
	&CLIP4Clip &2021
	    &45.7  &76.1 &83.3 &\textbf{2} &10.0 &49.3 &72.8 &78.6 &\textbf{1} &14.0  &405.1 \\
    &CLIP2Video &2021
	    &\textbf{46.1}  &76.0 &85.0 &\textbf{2} &9.7 &52.6 &75.5 &80.4 &\textbf{1} &9.7 &415.6\\
	&\textbf{CLIP-MSTDT (Ours)} &2021
	     &\textbf{46.1} &\textbf{76.5} &\textbf{85.4} &\textbf{2} &\textbf{9.6} &\textbf{55.3} &\textbf{81.8} &\textbf{89.7} &\textbf{1} &\textbf{6.3}  &\textbf{434.8}\\
	\hline 
	\multicolumn{1}{c}{\multirow{6}{*}{\rotatebox{270}{TGIF}}}
	&PVSE &2019
	    & 2.32 & 7.49 & 11.94 & 162 &- - & 2.17 &7.76 & 12.25 & 155 &- &-\\
	&HGR &2020
	    &4.17 &11.49 &16.53 &162 &-  &5.10 &14.02 &20.26 &104 &- &- \\
	&ARRN &2021
	    & 6.74 & 18.26 & 25.94 & 50 &-   & 5.35 &15.21 &21.84 &77 &- &- \\
	&CLIP4Clip &2021
	    &20.4 &39.6 &49.1 &\textbf{11} &128.5  
	    &29.5 &51.7 &61.7 &\textbf{5} &52.2 &252.0\\
	&CLIP2Video &2021
	    &20.5 &39.6 &48.8 &\textbf{11} &124.9 
	    &29.9 &52.8 &62.1 &\textbf{5} &49.6 &253.7\\
	&\textbf{CLIP-MSTDT (Ours)} &2021
	     &\textbf{21.0 } &\textbf{40.3} &\textbf{49.8} &\textbf{11} &\textbf{124.5} 
         &\textbf{30.4} &\textbf{53.0} &\textbf{62.8} &\textbf{5} &\textbf{48.4} &\textbf{257.3} \\
	\hline 
	\end{tabular}
	\caption{Experimental results on the MSR-VTT 9K/7K, MSVD and TGIF.}
	\label{tab:results_msrvtt9k}
\end{table*}

 \begin{figure*}[!tb]
    \centering
        \subfigure[Text-to-Video visualization]{
    \includegraphics[width = 1\textwidth]{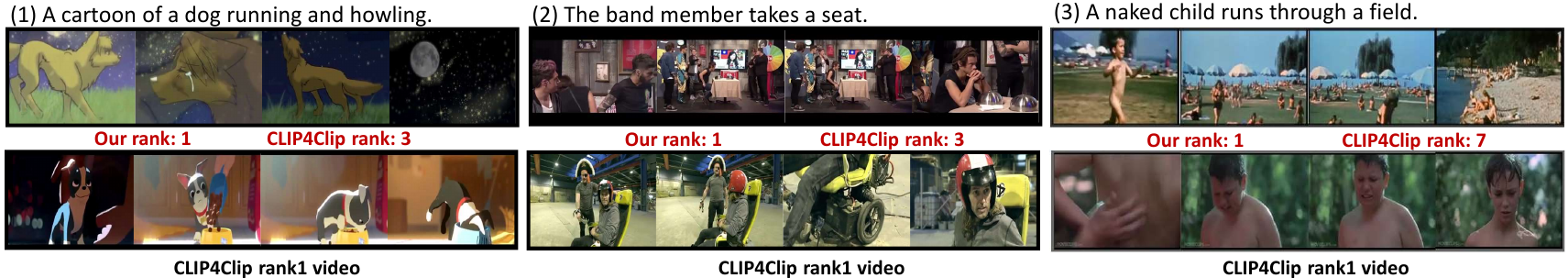} 
  }
  \subfigure[Video-to-Text visualization]{
     \includegraphics[width = 1\textwidth]{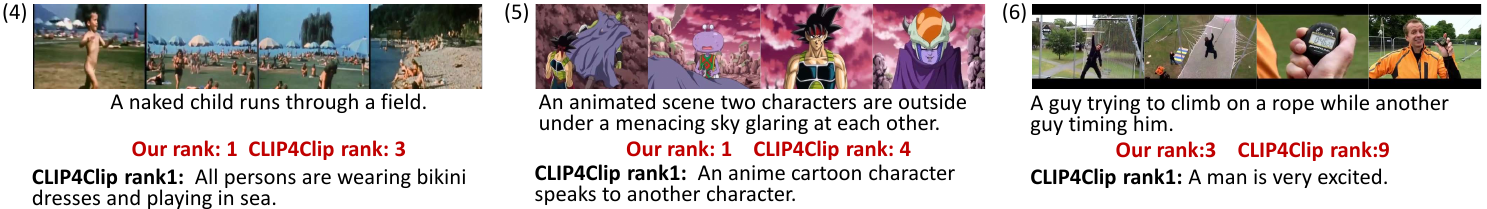}
}
    \caption{Visualization of Our method and CLIP4Clip on MST-VTT dataset.}
    \label{vis}
\end{figure*}

\begin{table} [!tb]
	\centering
	\small
	\begin{tabular}{l|ccccc}
	\hline
	\multicolumn{1}{c|}{\multirow{2}{*}{Method}} & \multicolumn{4}{c}{Text-to-Video} \\
	  & R@1 & R@5 & R@10 & MeanR \\
	  \hline
	Baseline &42.1 &70.4 &80.8 &16.2 \\
	+Bin  &43.2 &71.1 &81.8 &14.3  \\
    +LT  &42.8 &71.7 &81.6 &14.5 \\
    +MSTT  &44.2 &72.3 &83.2 &13.4 \\
    +MSTDT &44.6 &73.1 &82.8 &13.1  \\
	\hline
	Mean &44.6 &72.1 &81.9 &13.9 \\
	Concat &43.7  &72.4 &80.2 &14.0\\
	Attention &41.1  &69.2 &79.7 &17.8\\
	\hline
	scale=3 &44.0  &72.4 &81.7 &14.1\\
	scale=4 &44.3 &72.2 &83.3 &13.8\\
	scale=6 &44.2 &72.2 &81.4 &14.2\\
	scale=3,6 &44.6 &73.7 &82.4 &13.1\\
	scale=4,6 &44.4 &74.0 &82.8 &13.3\\
	\hline
	\end{tabular}
	\caption{Ablation studies. 
	}
	\label{ablation}
\end{table}

\begin{figure}[!tb]
    \centering
  \subfigure[]{
    \includegraphics[width =0.46\columnwidth]{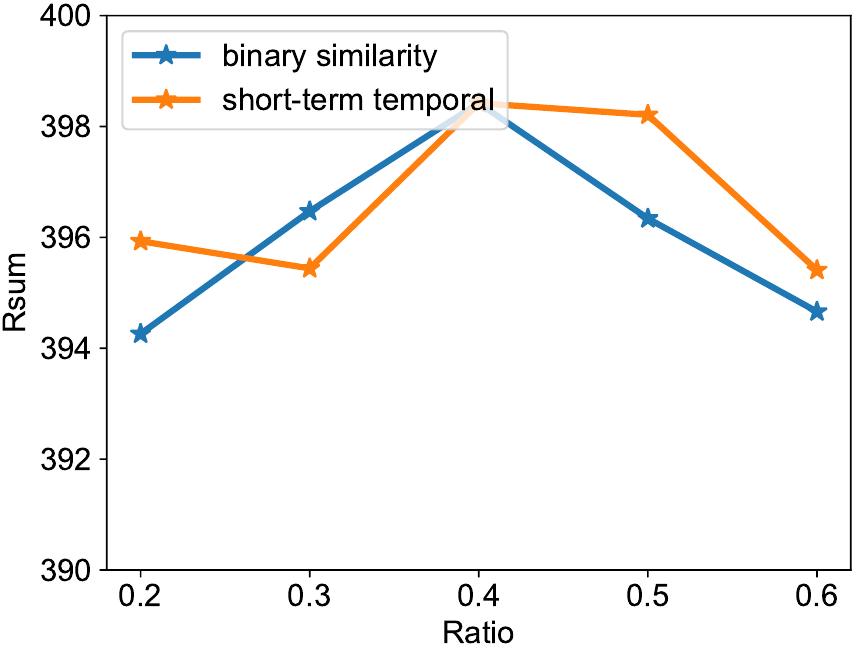}
}
    \subfigure[]{
    \includegraphics[width =0.48\columnwidth]{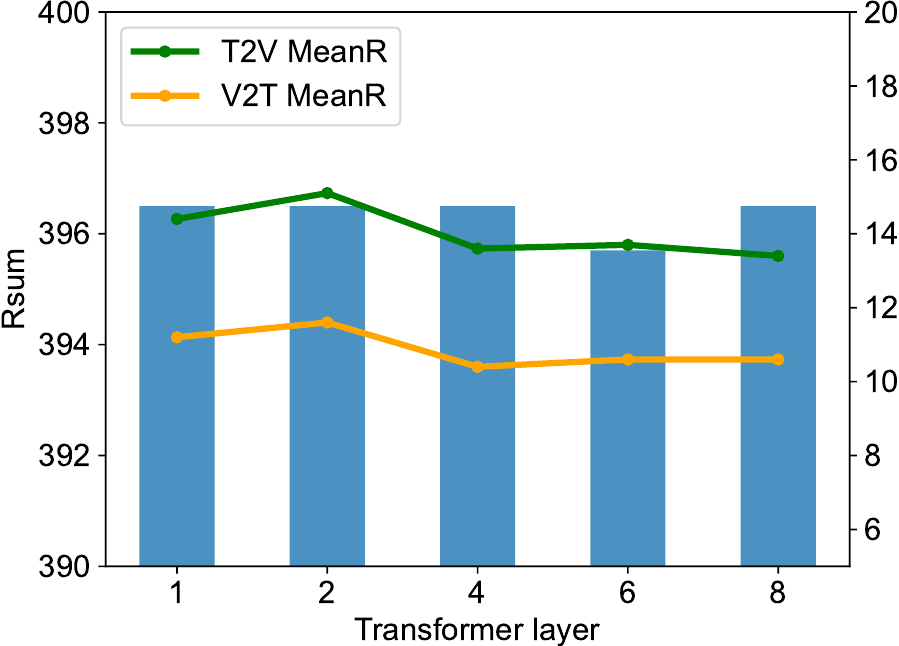}
}
    \caption{Comparison of different settings on trade-off rate (a) and transformer layers (b).}
    \label{ablation_fig}
\end{figure}

\subsection{Ablation Studies}

\subsubsection{How do different components in MSTDT framework contribute to its performance?}
Table \ref{ablation} illustrates the results of components ablation. ``Baseline" indicates directly using CLIP for fine-tuning with mean pooling fusion of frames. ``+Bin" indicates adding binary similarity loss, ``+LT" indicates using a long-term temporal transformer for video fusion. ``+MSTT" indicates using a short-term temporal transformer and a long-term temporal transformer for video fusion. ``+MSTDT" indicates using short-term temporal difference transformer and long-term temporal transformer for video fusion. In particular, the difference between ``+MSTT" and ``+MSTDT" is that the latter uses difference information and frame features, the former only uses frame features. Each component has brought the remarkable improvement. 

Binary similarity loss brings a significant improvement. It considers the distance of similar samples, so the similar samples always get higher scores. Besides, cross entropy loss can effectively ensure the score of ground truth pair, so the similar samples' scores could be high but smaller than ground truth pairs'.
Comparing baseline and "+LT", the latter performs better as the long-term temporal transformer acquires the temporal information of fine-tuning dataset. ``+MSTT" gains a larger gap than ``+LT". This phenomenon shows that considering short-term and long-term temporal simultaneously is essential. Though splitting neighbor frames into short clips, the model indeed works better. Besides, the comparison between ``+MSTT" and ``+MSTDT" shows that making use of difference information can also help the model perform better in temporal modeling.

\subsubsection{Which fusion strategies contribute to MSTDT's performance most?}
We analyze three fusion methods of the proposed multi-scale temporal difference transformer. As Table \ref{ablation} illustrated, ``Mean" is doing a mean pooling fusion on multi-scale features. ``Concat" is concatenating multi-scale features along the last dimension and mapping it into common space through a full connection layer. ``Attention" is doing a weighted sum on multi-scale features, we use the long-term transformer feature as query, multi-scale features as key and value. Among three fusion methods, ``Mean" performs best,  and ``Attention" performs worst. We think the mean pooling fusion could fairly and effectively integrate multi-scale temporal features. Attention fusion intends to pay more attention to the information that is similar to the long-term branch, so it would only cause redundancy of information and omission of other temporal information which not acquired by the long-term.

\subsubsection{Single-scale or Multi-scale?}
Table \ref{ablation} illustrates the various setting of scale list. It is obviously that multi-scale performs better than single-scale in general. Segmentation of different scales allows the sequence to generate multi-scale sub-sequences, and focusing on local information of different scales. Through mutli-scale, the model can learn local information at different levels.

\subsubsection{How do transformer layers affect MSTDT's performance?}
In Fig \ref{ablation_fig} (c), there is little fluctuation in $Rsum$ for different layers, but the value of $MeanR$ is getting lower with the increases of layers. When MSTDT has more transformer layers, it can learn more complex temporal relations. Limited by the data size, when the number of transformer layers exceeds 4, there will be no more significant improvement.
To further illustrate that the improvement of our model is not owing to the increase in the transformer layers. We compare the performance of 1-layer 3-scale MSTDT and 4-layer CLIP4Clip. These two models both contain 4 transformer layers in the temporal modeling module. As Table \ref{tab:results_msrvtt9k} shows, CLIP4Clip achieved 391.5 on MSTVTT 9K. Fig \ref{ablation_fig} (b) shows our model achieved more than 396. This result shows that our model performs better with the same number of transformer layers. It also shows that transformers with short-term information modeling capabilities could effectively improve performance. 

\subsubsection{How do trade-off rate affect MSTDT's performance?}
Fig \ref{ablation_fig} (a) illustrates the different setting of temporal rate $\alpha$, binary similarity loss rate $\beta$. 
When  $\alpha$ is between 0.4 and 0.5, it performs better. While the result drops when it exceeds 0.5. As static features dominate the information to be conveyed by the video, when $\alpha$ exceeds 0.5, static features will be ignored.
The main effect of binary similarity loss is to assist cross entropy and ensure the relative distance of similar samples. Similarly, when $\beta$ exceeds 0.4, our model will mainly emphasizes on relative distance, while ignoring the cross entropy. When $\beta$ is too small, it cannot promise the relative distance between similar samples.

\subsubsection{Qualitative results}
Figure \ref{vis} illustrates some retrieval results from MSR-VTT. For the text retrieval video part, the first line videos are ground truth. Then we provide the rank of our method and CLIP4Clip. The second-line videos are the result of CLIP4Clip. For the video retrieval text part, the first line captions are ground truth. We further give the retrieval rank of ground truth in our model and CLIP4Clip. The final line is the rank1 result for CLIP4Clip. 

We can see that our model does well when the video contains many actions and scene transitions. Additionally, our model can not only model static entities, such as ``dog", ``naked child", but also recognize the correctly distinguished actions like ``running", ``howling" and ``take a set". CLIP4Clip also recognizes those static entities and global action, such as ``playing", ``speaking". But it is limited to detailed actions. 
Example contains the action ``take a seat". CLIP4Clip just focuses on the state of ``sit", but our model pays attention to the process of ``take a seat". 
Our model successfully retrieved ground truth of example (4), while CLIP4Clip failed. The rank1 result of CLIP4clip shows that it ignored the scene transitions in the video.
Although two models both get wrong results in example (6), our model significantly improves the ranking of ground truth. 
These examples indicate our model can make CLIP more powerful by effectively making use of video temporal information, and doing a better job in video text retrieval.

\section{Conclusion}
\label{sec:conc}
We have proposed a novel method dubbed \textit{Multi-Scale Temporal Difference Transformer}, termed MSTDT, for video-text retrieval task. It consists of a short-term temporal difference transformer and a long-term temporal transformer. The former aims at modeling short-term dynamic information and increasing self-attention interaction capabilities , the latter mainly focuses on static information fusion and modeling long-term temporal information. The proposed binary similarity loss, which specialize distance of the similar samples also brings great improvement.
Comprehensive experiments and ablation studies on three datasets have demonstrated the superiority of our proposed method.

\bibliography{aaai22}
\end{document}